\documentclass[letterpaper]{article} % DO NOT CHANGE THIS
\usepackage{aaai2026}  % DO NOT CHANGE THIS
\usepackage{times}  % DO NOT CHANGE THIS
\usepackage{helvet}  % DO NOT CHANGE THIS
\usepackage{courier}  % DO NOT CHANGE THIS
\usepackage[hyphens]{url}  % DO NOT CHANGE THIS
\usepackage{graphicx} % DO NOT CHANGE THIS
\usepackage{natbib}  % DO NOT CHANGE THIS AND DO NOT ADD ANY OPTIONS TO IT
\usepackage{caption} % DO NOT CHANGE THIS AND DO NOT ADD ANY OPTIONS TO IT
\usepackage{algorithm}
\usepackage{algorithmic}

\usepackage{amsmath}
\usepackage{amsfonts}
\usepackage{bm} 

\usepackage{subcaption}
\usepackage{multirow}
\usepackage{pdfpages}

\newcommand{\Pre} {\mathrm{Pre}}
\newcommand{\Add} {\mathrm{Add}}
\newcommand{\Del} {\mathrm{Del}}
\newcommand{\pre} {\ensuremath{\mathit{pre}}}
\newcommand{\add} {\ensuremath{\mathit{add}}}
\newcommand{\del} {\ensuremath{\mathit{del}}}

\newcommand{\MSE} {\mathrm{MSE}}

\newcommand{\Equation} {Eq.}
\newcommand{\Figure} {Fig.}
\newcommand{\Table} {Tab.}
\newcommand{\Section}{Sec.}
\DeclareMathOperator*{\argmax}{\arg\!\max}

\newcommand{\mipvar}[1]{\text{#1}}

\usepackage{color}

\usepackage{newfloat}
\usepackage{listings}
\DeclareCaptionStyle{ruled}{labelfont=normalfont,labelsep=colon,strut=off} % DO NOT CHANGE THIS
\floatstyle{ruled}
\newfloat{listing}{tb}{lst}{}
\floatname{listing}{Listing}
\title{Learning Lifted Action Models from Unsupervised Visual Traces}
\author{
    Kai Xi\textsuperscript{\rm 1},
    Stephen Gould\textsuperscript{\rm 1},
    Sylvie Thi\'ebaux\textsuperscript{\rm 1, \rm 2}
}
\affiliations{
    \textsuperscript{\rm 1}School of Computing, The Australian National University\\
    \textsuperscript{\rm 2}LAAS-CNRS, Universit\'e de Toulouse\\
    oliver.xi@anu.edu.au, stephen.gould@anu.edu.au, sylvie.thiebaux@anu.edu.au
}

\begin{document}

\maketitle

\begin{abstract}
Efficient construction of models capturing the preconditions and effects of actions is essential for applying AI planning in real-world domains. Extensive prior work has explored learning such models from high-level descriptions of state and/or action sequences. In this paper, we tackle a more challenging setting: learning lifted action models from sequences of state images, without action observation. We propose a deep learning framework that jointly learns state prediction, action prediction, and a lifted action model. We also introduce a mixed-integer linear program (MILP) to prevent prediction collapse and self-reinforcing errors among predictions. The MILP takes the predicted states, actions, and action model over a subset of traces and solves for logically consistent states, actions, and action model that are as close as possible to the original predictions. Pseudo-labels extracted from the MILP solution are then used to guide further training. Experiments across multiple domains show that integrating MILP-based correction helps the model escape local optima and converge toward globally consistent solutions.
\end{abstract}

\begin{links}
    \link{Code}{https://github.com/xikaioliver/ROSAME}
    % \link{Datasets}{https://aaai.org/example/datasets}
    % \link{Extended version}{https://aaai.org/example/extended-version}
\end{links}

\section{Introduction}
AI planning enables intelligent agents to generate sequences of actions that achieve specified goals in complex environments. A core requirement for planning is an action model, specifying the preconditions and effects of actions, that allows a planner to reason about how actions change the state of the world. However, acquiring such models manually is often labor-intensive, error-prone, and expensive. This has motivated a growing body of research on learning action models from data, particularly from plan execution traces.

Much of the existing work focuses on learning symbolic action models from example traces composed of high-level, symbolic states and/or actions~\cite{ywj:07, cg:11, zk:13, aco:19, jls:21, ls:24, gjg:25}. While these approaches assume structured symbolic inputs, real-world data often comes in raw perceptual forms such as images or videos. With the advancement of deep learning, it has become feasible to make state and action predictions directly from such visual inputs, opening the door to learning action models from visual observations.

Two representative works in this direction are Latplan~\cite{akfm:22} and ROSAME~\cite{xgt:24}. Latplan is an unsupervised framework that learns action models from pairs of state images representing valid transitions in the environment. However, the resulting models are propositional and uninterpretable, with both states and actions encoded in an opaque latent space. To solve a new problem, Latplan requires images of the initial and goal states. In addition, its latent propositional representation is tied to a fixed object set and therefore cannot generalize to planning instances involving different objects. In contrast, ROSAME is a neuro-symbolic action model learner that computes probabilistic preconditions and effects for action schemas. When combined with deep learning-based state predictors, ROSAME learns lifted, human-readable action models from visual traces. However, it assumes that the executed actions are fully observed.

Here, we address some of the limitations of Latplan and ROSAME by learning lifted, human-readable action models from visual traces consisting of state images, without action supervision. To this end, we develop a deep learning framework that jointly learns to predict states, executed actions, and a lifted action model. While this framework can operate independently, we observe that it is often prone to prediction collapse and self-reinforcing errors among predictions, which prevent convergence to globally consistent solutions.

To overcome this, we introduce a mixed-integer linear program (MILP) as an external source of logical consistency. The MILP corrects a small subset of predicted traces and the action model so that they satisfy the logical constraints of planning while remaining close to the neural model’s original predictions. We extract pseudo-labels from the MILP solution and use them to supervise further training of the deep learning framework. This feedback loop enables the model to escape local optima and correct its own errors.

We evaluate our approach across several classical planning domains using two different types of visual representations. The experimental results show that integrating MILP-based correction substantially improves convergence and logical consistency in most cases, enabling recovery of the ground-truth action models without error.

\section{Preliminaries}
A typed \textbf{planning domain} $D= \langle \mathcal{T},\mathcal{P},\mathcal{A},M \rangle$ consists of: a set $\mathcal{T}$ of types, a set $\mathcal{P}$ of predicates, a set $\mathcal{A}$ of action schemas, and an action model $M$ specifying the preconditions, add and delete effects of each action schema.

The \textbf{types} in $\mathcal{T}$ are organized into a tree (or hierarchy). We say that a type $t'$ \textbf{\textit{subsumes}} type $t$ iff $t'$ is either $t$ or an ancestor of $t$ in the tree.
A \textbf{predicate} $\rho \in \mathcal{P}$ is a tuple $\langle name, params \rangle$, where $name$ is a symbol and $params$ is a list of types. We write $name(\rho)$ and $params(\rho)$ for the name and parameters of $\rho$, respectively.
Similarly, an \textbf{action schema} $\alpha \in \mathcal{A}$ is a tuple $\langle name, params \rangle$ whose name and parameters are denoted by $name(\alpha)$ and $params(\alpha)$, respectively. We borrow from SAM~\citep{jls:21} and define a \textbf{parameter binding} between a predicate $\rho$ and an action schema $\alpha$ as an \textit{injective} function $b_{\rho, \alpha}: params(\rho) \rightarrow params(\alpha)$ that maps every parameter $params(\rho)_i$ of $\rho$ to a parameter of $\alpha$ subsumed by $params(\rho)_i$. A \textbf{parameter-bound predicate} for an action schema $\alpha$ is a pair of the form $\langle \rho, b_{\rho, \alpha} \rangle$, where 
$b_{\rho,\alpha}$ is a valid parameter binding between $\rho$ and $\alpha$. 

Given $\mathcal{P}$ and $\mathcal{A}$, our goal is to learn an \textbf{action model} $M$ mapping each action schema $\alpha$, to a triple $M(\alpha)=\langle\Pre(\alpha), \Add(\alpha), \Del(\alpha)\rangle$ of sets of parameter-bound predicates representing its (positive) preconditions, add and delete effects. We denote by $\mathcal{M}(\mathcal{P}, \mathcal{A})$ the set of candidate action models over the given predicates and action schemas.

A \textbf{planning instance} $I=\langle O, D\rangle$ consists of a set of objects $O$ and a planning domain $D$. Each \textbf{object} $o\in O$ is associated with a leaf type $\text{type}(o) \in \mathcal{T}$ of the type hierarchy. A \textbf{binding} of a predicate $\rho$ is an \textit{injective} function $b_\rho: params(\rho) \rightarrow O$ mapping every parameter $params(\rho)_i$ of $\rho$ to an object $o \in O$ such that $params(\rho)_i$ subsumes $type(o)$. A \textbf{proposition} $p$ is a pair $\langle \rho, b_\rho \rangle$ where $\rho$ is a predicate and $b_\rho$ is a binding for $\rho$.  We say that proposition $p = \langle \rho, b_\rho \rangle$ is a ground instance of predicate $\rho$. Similarly, a binding of an action schema $\alpha$ is defined like a binding of a predicate, i.e., an \textit{injective} function $b_\alpha: params(\alpha) \rightarrow O$ where $\forall i$, $params(\alpha)_i$ subsumes $type(b_\alpha(params(\alpha)_i))$ . An \textbf{action} $a = \langle \alpha, b_\alpha\rangle$, where $\alpha$ is an action schema and $b_\alpha$ is a binding for $\alpha$, is a ground instance of the action schema $\alpha$. The preconditions of a grounded action $a = \langle \alpha, b_\alpha \rangle$, denoted by $\Pre(a)$, is the set of propositions created by taking every parameter-bound predicate $\langle \rho, b_{\rho, \alpha} \rangle$ in the preconditions of the action schema $\Pre(\alpha)$ and grounding $\rho$ with the binding $b_\alpha \circ b_{\rho, \alpha}$. The effects of a grounded action $a$ are defined in a similar manner, and the action model $M(a)$ for the grounded action $a$ is $\langle \Pre(a), \Add(a), \Del(a) \rangle$. Given a planning instance $I$, we write $P_{I}$ for the set of propositions, $A_{I}$ for the set of actions, and $S \subseteq 2^{P_{I}}$ for the set of \textbf{states}.

Let $I$ be a planning instance, $s\in S$ be a state, $a\in A_{I}$ be an action. We say that $a$ is \textbf{applicable} in $s$ iff $\Pre(a) \subseteq s$. The result of applying $a$ in $s$ is the \textbf{successor} state $res(s, a) = (s \setminus \Del(a)) \cup \Add(a)$. A \textbf{state trace} for planning instance $I$ is a state sequence $e = (s_1, s_2, \ldots, s_{|e|+1})$. The \textbf{length} $|e|$ of the trace is its number of transitions. We say that a trace is \emph{$k$-step} if it has length $k$. We refer to $s_1$ as the \textbf{initial state} and $s_{|e|+1}$ as the \textbf{final state} of the trace. State trace $e$ is \textbf{\textit{consistent}} with an action model $M$ if and only if, according to $M$, for all $t \in 1, \ldots, |e|$, there exists an action $a_t$ that is applicable in $s_t$ and $res(s_t, a_t) = s_{t+1}$. In the following, we write $\mathcal{E}_M^k$ for the set of state traces of length $k$ that are consistent with action model $M$.

\begin{figure}[t]
  \centering
  \includegraphics[width=0.65\linewidth]{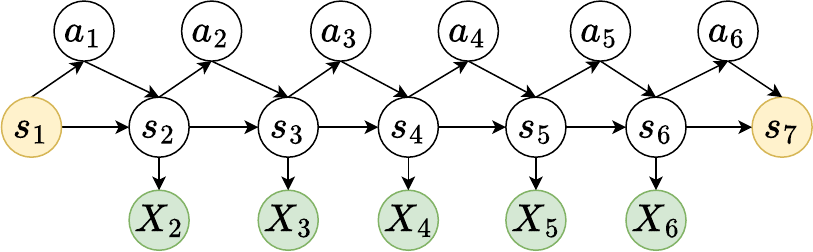}
  \caption{Observations in a visual trace $\tau$ compared to the state trace $e$. 
  Symbol $s$ denotes states and $X$ denotes images. Shaded nodes are observed.}
  \label{fig:trace illustration}
\end{figure}

\begin{figure}[t]
  \centering
  \includegraphics[width=\linewidth]{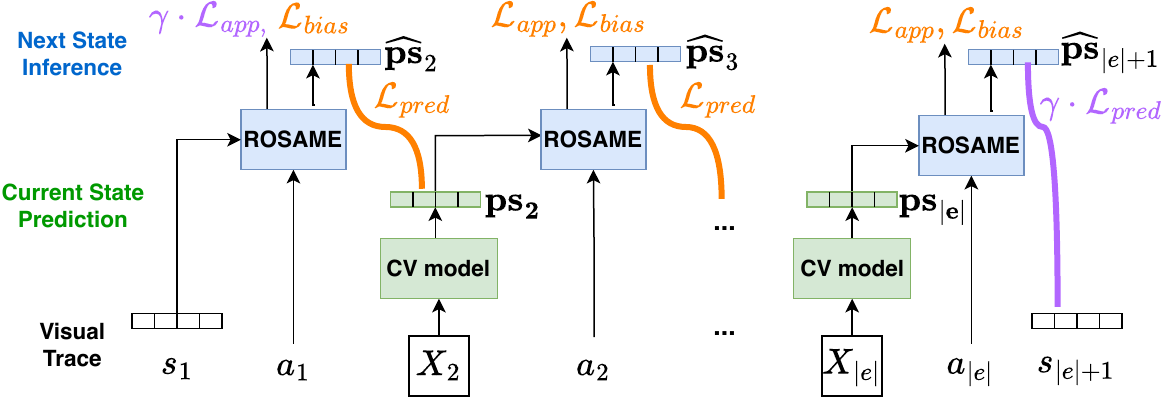}
  \caption{Integrating ROSAME with a state predictor to learn action models 
  from visual traces with fully observed actions, adapted from ~\citet{xgt:24}.}
  \label{fig:rosame-I}
\end{figure}

\section{Problem Formulation}
\label{sec: problem formulation}
We now formalize the problem considered in this paper. For an execution trace $e$, we observe only a \textbf{visual trace},
\[
    \tau = (s_1, X_2, X_3, \ldots, X_{|e|}, s_{|e|+1}),
\]
where $s_1$ and $s_{|e|+1}$ are states, and each $X_t$ is an image of the corresponding intermediate state. We are given a dataset $\{\tau_j\}_{j=1}^{n}$ of such visual traces generated under a domain $D^M=\langle \mathcal{T}, \mathcal{P}, \mathcal{A}, M \rangle$ for an unknown action model $M \in \mathcal{M}(\mathcal{P}, \mathcal{A})$. Our assumptions are as follows.
\begin{itemize}
\item The initial and final states are fully observed. Having known initial and final states helps prevent degenerate solutions and ensures that the learned state representation remains human-readable. Importantly, we do not include images for these states.
\item Each intermediate observation $X_t$ is a fully observed image of its underlying state, i.e., it fully determines the state. Note that the converse is not true: several different images can represent the same state.
\item The predicate set $\mathcal{P}$ and action schema set $\mathcal{A}$ (names and parameter signatures) are given.
\item For each trace, the object set $O$ is given. For simplicity, our notations assume that all traces share the same object set $O$, but in principle, traces may come from instances with different object sets.
\item Our implementation uses ROSAME~\cite{xgt:24} as a component, where add effects and preconditions cannot intersect, and only preconditions may be deleted. This assumption can easily be relaxed.
\end{itemize}
\Figure~\ref{fig:trace illustration} illustrates our observations in a 6-step visual trace compared to the ground-truth state trace.

Given a visual trace $\tau_j$, we wish to use a neural network with parameters $\theta$ to predict the underlying state trace $e$. Ideally, we aim to jointly choose an action model $M$ and neural network parameters $\theta$ to maximize the log-likelihood of the visual traces being observations of state traces consistent with the action model $M$:
\begin{align*}
    \ell(M, \theta) = \sum_{j=1}^{n} \log\!\left( \sum_{e \in {\cal E}_M^{|\tau_j|}} \Pr(e \mid \tau_j; \theta) \right).
\end{align*}
In practice, this formulation is intractable: evaluating joint distributions over all propositions across all time steps and summing over $e \in \mathcal{E}_M^{|\tau_j|}$ is computationally prohibitive. Furthermore, optimizing over discrete action models is hard.

To make learning feasible, we introduce the following three relaxations:
\begin{enumerate}
    \item We relax the trace prediction to be the product of marginal state probabilities at each time step.
    \item Following ROSAME, we further relax each state prediction to a \textbf{probabilistic state vector} $\bm{ps}_t \in [0,1]^{|P_I|}$ listing the marginal probabilities of each proposition.
    \item Also following ROSAME, instead of optimizing directly over the discrete space of action models, we compute a probability distribution over action models and modify the successor operator $res$ to be probabilistic so that $res(\bm{ps}_t, a)$ gives the expected next probabilistic state.
\end{enumerate}

Under these relaxations, the probability of a trace $e$ given a visual trace $\tau$ factorizes as
\begin{align*}
    \Pr(e \mid \tau) 
    &= \prod_t \Pr(s_{t+1} \mid s_t, X_{t+1}) \\
    &= \prod_t \sum_{a \in A_I} \Pr(s_{t+1}, a \mid s_t, X_{t+1}) \\
    &= \prod_t \sum_{a \in A_I} \Pr(a \mid s_t, s_{t+1})\Pr(s_{t+1} \mid s_t, X_{t+1}),
\end{align*}
where $X_t$ denotes the image observation at time $t$. Since the state images are fully observed, $\Pr(s_{t+1} \mid s_t, X_{t+1}) = \Pr(s_{t+1} \mid X_{t+1})$, giving
\begin{align*}
    \Pr(e \mid \tau)
    = \prod_{t} \sum_{a \in A_I}
      \Pr(a \mid s_t, s_{t+1})\,
      \Pr(s_{t+1} \mid X_{t+1}).
\end{align*}

We predict these distributions with two neural networks: a state predictor with parameters $\theta_s$ that estimates $\Pr(s_t \mid X_t; \theta_s)$ and outputs $\bm{ps}_t$, and an action predictor with parameters $\theta_a$ that estimates $\Pr(a \mid s_t, s_{t+1}; \theta_a)$. The initial and final states come directly from the input; intermediate $\bm{ps}_t$ are predicted from images.

We compute a probability distribution over action models with parameters
$\theta_M$, which can be obtained using ROSAME (see \Section~\ref{sec: ROSAME}).
We then minimize
\begin{align*}
    \mathop{\mathbb{E}}_{M \sim \Pr(M;\theta_M)} \sum_{j,t} \mathcal{L}_{j,t}(M).
\end{align*}
where for each trace $j$ and step $t$, the loss $\mathcal{L}_{j,t}(M)$ is
\begin{align*}
    \mathop{\mathbb{E}}_{a \sim \Pr(a \mid \bm{ps}_t, \bm{ps}_{t+1}; \theta_a)}
    \Big[
      \|res(\bm{ps}_t, a) - \bm{ps}_{t+1}\|_2^2
      + \mathcal{L}(a, \bm{ps}_t)
    \Big],
\end{align*}
\normalsize
and $\mathcal{L}(a, \bm{ps}_t)$ is a relaxed loss measuring applicability of action $a$ in state $\bm{ps}_t$.

Note that the loss decomposes over traces and time steps. For clarity in what follows, we define the loss for a single time step~$t$, omitting the trace index~$j$; the final objective is the sum of these losses across all traces and time steps.

\section{ROSAME}
\label{sec: ROSAME}
\citet{xgt:24} presented a neuro-symbolic action model learner—ROSAME—and integrated it with a deep learning state predictor to enable action model learning from sequences of interleaved state images and observed actions. We build on this work by extending ROSAME to enable action model learning without observing the actions.

ROSAME is a learnable model that takes as input a grounded action and outputs the grounded preconditions, add effects, and delete effects of that action. To achieve this, ROSAME learns a neural network for each action schema, whose outputs are the lifted preconditions and effects for the schema. Given an action $a_t$ performed at step $t$, ROSAME first looks up its action schema $\alpha$ and computes three vectors: $\bm{pre}_{\alpha}$, $\bm{add}_{\alpha}$, and $\bm{del}_{\alpha}$, where each component represents the probability of a parameter-bound predicate being a precondition, an add effect, or a delete effect, respectively. These lifted conditions and effects are then mapped to their grounded counterparts for that action, i.e., $\bm{pre}_{a_t}$, $\bm{add}_{a_t}$, and $\bm{del}_{a_t}$. Each value in these grounded vectors represents the probability that a proposition is a precondition or effect of the action. This probability is copied from the corresponding lifted vector if a parameter binding exists that unifies the proposition with the action; otherwise, it is set to zero.

Given a state prediction $\bm{ps}_t$ and an observed action $a_t$, ROSAME computes the preconditions and effects of $a_t$ and infers the next state as
\begin{align}
    \widehat{\bm{ps}}_{a_t} = \bm{ps}_t \odot (1-\bm{\del}_{a_t}) + (1-\bm{ps}_t) \odot \bm{\add}_{a_t},
\end{align}
where $\odot$ denotes elementwise product.

This predicted state should be close to the next state prediction $\bm{ps}_{t+1}$ derived from the next state image. Accordingly, we define the prediction loss as:
\begin{align}
\label{equation: loss_pred}
\mathcal{L}_{\text{pred}} = \MSE(\widehat{\bm{ps}}_{a_t}, \bm{ps}_{t+1})
\end{align}

In addition, all preconditions of $a_t$ must hold in $\bm{ps}_t$. That is, a proposition cannot be a precondition of $a_t$ and simultaneously be false in $\bm{ps}_t$. This gives rise to an action applicability loss:
\begin{align}
\label{equation: loss_app}
    \mathcal{L}_{\text{app}} = \MSE(\bm{\pre}_{a_t} \odot (\mathbf{1}-\bm{ps}_t), \mathbf{0})
\end{align}

ROSAME also introduces a prior bias loss that encourages learning parameter-bound predicates as preconditions, in order to recover prevail conditions—preconditions that are not deleted by actions~\cite{bcn:95}:
\begin{align}
\label{equation: loss_bias}
    \mathcal{L}_{\text{bias}} = \MSE(\bm{pre}_{a_t}, \mathbf{1})
\end{align}
If a proposition remains true both before and after an action, it could be a prevail condition, an add effect, or irrelevant to the action. In such cases, the prior bias loss encourages including the proposition as a precondition.

The total loss used to train ROSAME for the observed action $a_t$ is:
\begin{align}
\label{equation: ROSAME_loss}
    \mathcal{L}_{a_t} = \mathcal{L}_{\text{pred}} + \mathcal{L}_{\text{app}} + \lambda \cdot \mathcal{L}_{\text{bias}}
\end{align}
where $\lambda$ is a hyperparameter that controls the strength of the bias term. Gradients are directly backpropagated through the neural networks associated with the action schemas, allowing the model to learn a consistent lifted action model.

For different actions, ROSAME operates like a lookup table: it selects and runs the neural network corresponding to the action schema. ROSAME provides a differentiable and logically consistent way to compute preconditions and effects. In principle, it can be integrated with any deep learning framework that expects such structured inputs and outputs. \Figure~\ref{fig:rosame-I} illustrates how ROSAME is integrated with a deep learning-based state predictor to enable action model learning from sequences of state images and fully observable actions. A hyperparameter $\gamma \geq 1$ is used to upweight the prediction loss at the final step, since the final state is fully observed.

We treat ROSAME as a black-box action model learner. In particular, it provides the probability distribution over action models $\Pr(M; \theta_M)$ used in our framework. Conceptually, our method does not rely on any ROSAME-specific architecture; it is compatible with any action model learner that produces probabilistic preconditions and effects in a similar form.

\section{Deep Learning Framework}
\label{sec: dl_framework}
\textbf{Action Prediction and Regularization.}
In this work, we assume that actions are not observed in the visual traces. As such, we introduce an additional neural network that predicts the action from two consecutive states. Its output is a probability distribution over actions, $\Pr(a_t \mid \bm{ps}_t, \bm{ps}_{t+1})$.

To compute the loss in the absence of observed actions, we take the expected value over the ROSAME loss in \Equation~\ref{equation: ROSAME_loss} for all possible actions to get:
\begin{align}
\label{equation: loss_dl}
    \bar{\mathcal{L}}_{a_t} = \sum_{a_t \in A_I} \Pr(a_t \mid \bm{ps}_t, \bm{ps}_{t+1}) \mathcal{L}_{a_t}
\end{align}
This formulation allows the preconditions and effects for different actions to remain separate during next-state inference and applicability checking (\Equation~\ref{equation: loss_pred} and~\ref{equation: loss_app}), while also providing a training signal that encourages the neural network to predict the most consistent action for each step.

However, in practice, action prediction often gets stuck in local minima under the above loss. We found that this arises because some actions are much more similar to each other—both visually and semantically—than to others. For example, in blocksworld, the only difference between $\mathrm{pickup}(b1)$ and $\mathrm{unstack}(b1, b2)$ lies in the state they apply to. Confusing them incurs less loss than predicting an unrelated action like $\mathrm{putdown}(b1)$. These ambiguities lead to a loss landscape with many local minima, making it difficult for gradient-based optimization to reach the correct solution.

To mitigate this issue, we introduce a simple self-regularization loss that helps the learning process escape such local minima by using the locally best action as an auxiliary target. For an action $a_t$, the log-probability that its inferred next state $\widehat{\bm{ps}}_{a_t}$ matches the state prediction $\bm{ps}_{t+1}$ is:

{\small
\begin{align}
     \!\!\!\!\log \mathop{\Pr}\nolimits_{\text{pred}} \!= \!\sum_{i=1}^{|P_I|} \!\log \!\left[ \widehat{\bm{ps}}_{a_t} \!\!\odot \!\bm{ps}_{t+1} \!+ \!(1\!-\!\widehat{\bm{ps}}_{a_t})\!\odot\! (1\!-\!\bm{ps}_{t+1})\!\right]_i
\end{align}
}%
\normalsize
\noindent Similarly, the log-probability of action $a_t$ being applicable in state $\bm{ps}_t$ is:
\begin{align}
     \log \mathop{\Pr}\nolimits_{\text{app}} = \sum_{i=1}^{|P_I|} \log \left[\bm{1} - \bm{\pre}_{a_t} \odot (\mathbf{1}-\bm{ps}_t)\right]_i
\end{align}
We define the locally best action as:
\begin{align}
    a_t^\ast = \argmax \left[ \log \mathop{\Pr}\nolimits_{\text{pred}} + \log \mathop{\Pr}\nolimits_{\text{app}} \right]
\end{align}
Finally, we augment the loss from \Equation~\ref{equation: loss_dl} with an additional cross-entropy term that encourages the action prediction to match $a_t^\ast$:
\begin{align}
    \mathcal{L}_t = \bar{\mathcal{L}}_{a_t} + \mathrm{CE}\!\left(\Pr(a_t \mid \bm{ps}_t, \bm{ps}_{t+1}),\; a_t^\ast\right)
\end{align}

\begin{figure}[t]
  \centering
  \includegraphics[width=\linewidth]{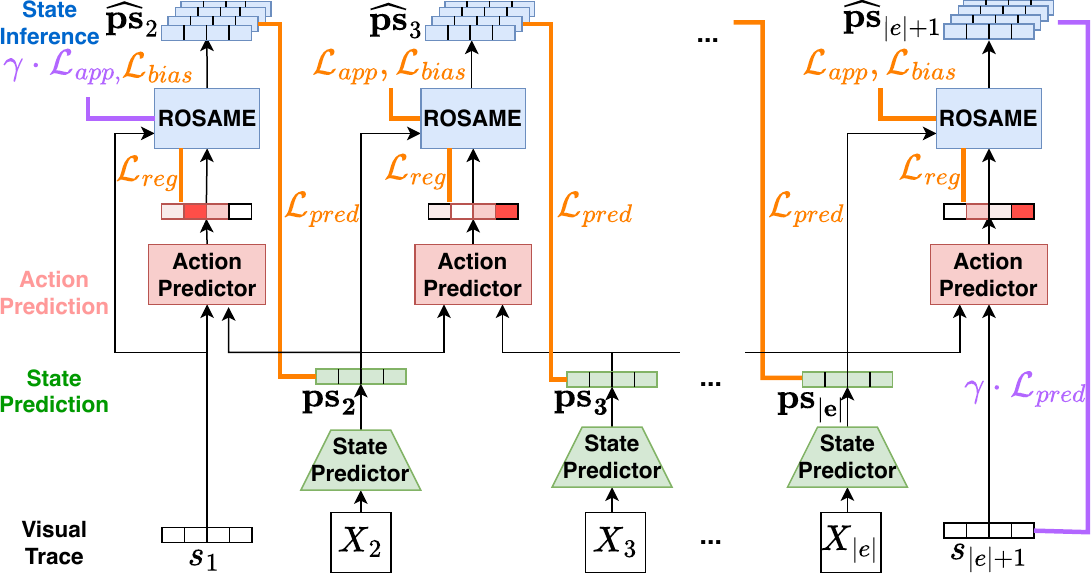}
  \caption{
    Deep learning framework that predicts states, actions and the action model; a scaling factor $\gamma \geq 1$ emphasizes losses at the fully observed initial and final steps.
  }
  \label{fig:DL framework}
\end{figure}

\Figure~\ref{fig:DL framework} presents our framework for learning a lifted action model from visual traces. Since we assume that the initial and final states are fully observed, we introduce a hyperparameter $\gamma \geq 1$ to scale the first-step applicability loss and the last-step prediction loss. This encourages consistency with the available observations.

\section{MILP Solution Fixer}
\label{sec:MIP}
The framework described above can, in principle, solve the relaxed task defined in \Section~\ref{sec: problem formulation}. A globally optimal solution exists in which the predicted states, actions, and action model align with the ground truth. However, in practice, without any action supervision, the system is susceptible to collapse to degenerate solutions—internally consistent predictions of states, actions, and an action model that do not reflect the true underlying dynamics. Moreover, small systematic mistakes early in training can lead to self-reinforcing local optima, where errors in the state predictions, action predictions, and the learned action model repeatedly support one another rather than being corrected. In such cases, the model can become stuck in configurations where significant inconsistencies with the visual traces persist, and gradient-based training alone struggles to repair these errors.

To prevent such collapse and correct persistent logical inconsistencies, we introduce an external source of regularization. Specifically, we formulate a mixed-integer linear programming (MILP) problem that extracts a logically consistent set of states, actions, and a lifted action model that remain as close as possible to the outputs of the deep learning framework. In the deep learning framework, logical constraints are imposed only through soft relaxations and can therefore be violated. The MILP repairs these errors by enforcing hard constraints while making minimal edits to the predicted states, actions, and action model. This corrective mechanism guides learning away from degenerate solutions and enables the system to escape self-reinforcing local optima. Training converges when neural predictions and MILP solutions agree globally across all training examples.

In the MILP, we define binary decision variables \mipvar{pre}, \mipvar{add}, and \mipvar{del}, indexed by action schema $\alpha$ and parameter-bound predicate $\langle \rho, b_{\rho, \alpha} \rangle$, indicating whether $\langle \rho, b_{\rho, \alpha} \rangle$ is a lifted precondition, add effect, or delete effect of $\alpha$. These correspond to the structure produced by ROSAME. We also define binary variables \mipvar{act} and \mipvar{hol}, where \mipvar{act}\([a,t]\) indicates whether action $a$ is executed at time $t$, and \mipvar{hol}\([p,t]\) indicates whether proposition~$p$ holds at time~$t$.

We further introduce auxiliary step-level indicators \!\mipvar{stepadd}, \!\mipvar{stepdel}, and \!\mipvar{steppre}, where \mipvar{stepadd}\([p,t]\), \!\mipvar{stepdel}\([p,t]\), and \!\mipvar{steppre}\([p,t]\)\! indicate whether proposition $p$ is added, deleted or required as a precondition at time~$t$, respectively.

For a proposition $p=\langle \rho, b_\rho \rangle$, we define its \textbf{actors} as the set of all actions whose parameters can be bound to instantiate $p$: $\text{actors}(p) = \{a=\langle \alpha, b_\alpha \rangle \mid \exists b_{\rho,\alpha} \text{ s.t. } b_\rho = b_\alpha \circ b_{\rho, \alpha}\}$.

The inputs to the MILP are the state and action predictions from the deep learning model, the fully observed initial and final states, and the lifted action model predicted by ROSAME. Specifically, \mipvar{OBS}\((p,t)\) denotes the predicted probability that proposition $p$ holds at time $t$, and \mipvar{OBS}\((a,t)\) denotes the probability that action $a$ is executed at time~$t$. Similarly, $\text{\mipvar{ROS}}_{\text{\mipvar{Pre}}}$, $\text{\mipvar{ROS}}_{\text{\mipvar{Add}}}$, and $\text{\mipvar{ROS}}_{\text{\mipvar{Del}}}$ denote the predicted probabilities that a parameter-bound predicate is a precondition, add effect, or delete effect of an action schema.

The objective of the MILP is to maximize the \textbf{agreement} between the binary decision variables and the soft predictions. For example, the agreement term for propositions is:
{\small
\begin{align*}
    \mipvar{OBS}(p, t) \cdot \mipvar{hol}[p, t] + (1 - \mipvar{OBS}(p, t)) \cdot (1 - \mipvar{hol}[p, t]),
\end{align*}
}%
which is rewritten in linear form below (up to a constant).

We present the formulation for a single trace below; extending it to multiple traces simply replicates the \mipvar{hol}, \mipvar{act}, and step-level indicator variables for each trace, while sharing the lifted action model variables \mipvar{pre}, \mipvar{add}, and \mipvar{del} across all traces. The objectives in \Equation~\ref{eq: objective_state_agree}–\ref{eq: objective_action_agree} are then summed over all traces. \Equation~\ref{eq: mip_prior} adds a prior bias term consistent with \Equation~\ref{equation: ROSAME_loss}, using the same hyperparameter~$\lambda$.

\allowdisplaybreaks
{\small
\begin{align}
\mathop{\text{maximize}}_{\mipvar{pre}, \mipvar{add}, \mipvar{del}, \mipvar{act}, \mipvar{hol}}
\label{eq: objective_state_agree}
\sum_{p, t} \left(2 \cdot \mipvar{OBS}(p, t) - 1\right) \cdot \mipvar{hol}[p, t] \quad +\\[-1ex]
\label{eq: objective_action_agree}
\quad \sum_{a, t} \left(2 \cdot \mipvar{OBS}(a, t) - 1\right) \cdot \mipvar{act}[a, t] \quad + \\[-1ex]
\!\!\!\!\sum_{\alpha, \langle \rho, b_{\rho, \alpha} \rangle} \!\!\!\!\!\!\left(2 \!\cdot \!\mipvar{ROS}_{\mipvar{Pre}}(\alpha, \langle \rho, b_{\rho, \alpha} \rangle) \!- \!1\right) \!\cdot \!\mipvar{pre}[\alpha, \langle \rho, b_{\rho, \alpha} \rangle]+ \label{eq: rosame_pre} \\[-1ex]
\!\!\!\!\sum_{\alpha, \langle \rho, b_{\rho, \alpha} \rangle} \!\!\!\!\!\!\left(2 \!\cdot \!\mipvar{ROS}_{\mipvar{Add}}(\alpha, \langle \rho, b_{\rho, \alpha} \rangle) \!- \!1\right) \!\cdot \!\mipvar{add}[\alpha, \langle \rho, b_{\rho, \alpha} \rangle]+ \label{eq: rosame_add} \\[-1ex]
\!\!\!\!\sum_{\alpha, \langle \rho, b_{\rho, \alpha} \rangle} \!\!\!\!\!\!\left(2 \!\cdot\! \mipvar{ROS}_{\mipvar{Del}}(\alpha, \langle \rho, b_{\rho, \alpha} \rangle) \!-\! 1\right) \!\cdot\! \mipvar{del}[\alpha, \langle \rho, b_{\rho, \alpha} \rangle]+ \label{eq: rosame_del} \\[-1ex]
\!\!\!\!\sum_{\alpha, \langle \rho, b_{\rho, \alpha} \rangle}
\lambda \cdot \mipvar{pre}[\alpha, \langle \rho, b_{\rho, \alpha} \rangle]\label{eq: mip_prior} ~~
\end{align}}\\[-6ex]
% Domain constraints
\allowdisplaybreaks
{\small
\begin{align}
% Domain constraints
\text{s.t.} \quad & \forall \alpha \text{ and} \langle\rho, b_{\rho, \alpha}\rangle, \notag \\
\label{eq: add_pre_not_intersect}
&\quad\quad \mipvar{pre}[\alpha, \langle \rho, b_{\rho, \alpha} \rangle] + \mipvar{add}[\alpha, \langle \rho, b_{\rho, \alpha} \rangle] \leq 1 \\
\label{eq: del_must_be_pre}
&\quad\quad \mipvar{pre}[\alpha, \langle \rho, b_{\rho, \alpha} \rangle] \geq \mipvar{del}[\alpha, \langle \rho, b_{\rho, \alpha} \rangle] \\
\label{eq: init-constraints_true}
&\forall p \in s_{1}, \quad\quad \hphantom{||+} \mipvar{hol}[p, 1] = 1 \\
\label{eq: init-constraints_false}
&\forall p \notin s_{1}, \quad\quad  \hphantom{||+} \mipvar{hol}[p, 1] = 0\\
\label{eq: goal-constraints_true}
&\forall p \in s_{|e|+1}, \quad\quad \mipvar{hol}[p, |e|+1] = 1 \\
\label{eq: goal-constraints_false}
&\forall p \notin s_{|e|+1}, \quad\quad \mipvar{hol}[p, |e|+1] = 0 \\
% Unique action per step
\label{eq: unique-action}
&\forall t, \quad\quad \sum_{a \in A_I} \mipvar{act}[a, t] = 1 \\
% Step-level min effects
&\forall t, \forall p = \langle \rho, b_\rho \rangle, \forall a = \langle \alpha, b_\alpha \rangle \in \text{actors}(p), \notag \\
\label{eq: step_on_begin}
&\quad\quad \mipvar{stepadd}[p, t] \leq \mipvar{add}[\alpha, \langle \rho, b_{\rho, \alpha} \rangle] + 1 - \mipvar{act}[a, t] \\
&\quad\quad \mipvar{stepadd}[p, t] \geq \mipvar{add}[\alpha, \langle \rho, b_{\rho, \alpha} \rangle] + \mipvar{act}[a, t] - 1 \\
&\quad\quad \mipvar{stepdel}[p, t] \leq \mipvar{del}[\alpha, \langle \rho, b_{\rho, \alpha} \rangle] + 1 - \mipvar{act}[a, t] \\
&\quad\quad \mipvar{stepdel}[p, t] \geq \mipvar{del}[\alpha, \langle \rho, b_{\rho, \alpha} \rangle] + \mipvar{act}[a, t] - 1 \\
&\quad\quad \mipvar{steppre}[p, t] \leq \mipvar{pre}[\alpha, \langle \rho, b_{\rho, \alpha} \rangle] + 1 - \mipvar{act}[a, t] \\
\label{eq: step_on_end}
&\quad\quad \mipvar{steppre}[p, t] \geq \mipvar{pre}[\alpha, \langle \rho, b_{\rho, \alpha} \rangle] + \mipvar{act}[a, t] - 1 \\
% Step-level max effects
\label{eq: step_off_begin}
&\quad\quad \mipvar{stepadd}[p, t] \leq \sum_{a \in \text{actors}(p)} \mipvar{act}[a, t] \\
&\quad\quad \mipvar{stepdel}[p, t] \leq \sum_{a \in \text{actors}(p)} \mipvar{act}[a, t] \\
\label{eq: step_off_end}
&\quad\quad \mipvar{steppre}[p, t] \leq \sum_{a \in \text{actors}(p)} \mipvar{act}[a, t] \\
% Frame effects
\label{eq: stepadd_next_hol}
&\quad\quad \mipvar{hol}[p, t+1] \geq \mipvar{stepadd}[p, t] \\
\label{eq: stepdel_not_next_hol}
&\quad\quad 1 - \mipvar{hol}[p, t+1] \geq \mipvar{stepdel}[p, t] \\
\label{eq: steppre_hol}
&\quad\quad \mipvar{hol}[p, t] \geq \mipvar{steppre}[p, t] \\
% Frame explanation
\label{eq: frame_add}
&\quad\quad \mipvar{stepadd}[p, t] \geq \mipvar{hol}[p, t+1] - \mipvar{hol}[p, t] \\
\label{eq: frame_del}
&\quad\quad \mipvar{stepdel}[p, t] \geq \mipvar{hol}[p, t] - \mipvar{hol}[p, t+1]
\end{align}}%

\Equation~\ref{eq: add_pre_not_intersect} and \ref{eq: del_must_be_pre} enforce the validity of the lifted action model according to ROSAME’s assumptions: add effects and preconditions cannot intersect, and only preconditions may be deleted.
\Equation~\ref{eq: init-constraints_true}–\ref{eq: goal-constraints_false} enforce that the proposition truth values at time steps $1$ and $|e|+1$ match the observed initial and final states.
\Equation~\ref{eq: unique-action} ensures that exactly one action is executed per timestep.

\Equation~\ref{eq: step_on_begin}–\ref{eq: step_on_end} use variables \mipvar{act}\([a,t]\) to activate the auxiliary step-level indicators when an action is executed. These constraints ensure that, if action $a$ is chosen at time~$t$, its corresponding preconditions and effects are correctly reflected in \mipvar{steppre}, \mipvar{stepadd}, and \mipvar{stepdel}. Conversely, \Equation~\ref{eq: step_off_begin}–\ref{eq: step_off_end} ensure that the auxiliary step-level indicators remain zero if no action capable of influencing proposition $p$ is executed at $t$.

Finally, \Equation~\ref{eq: stepadd_next_hol}–\ref{eq: steppre_hol} ensure that the step-level effects and preconditions are properly reflected in the pre- and post-states, and \Equation~\ref{eq: frame_add}–\ref{eq: frame_del} encode the frame axioms.

In addition to probabilistic predictions encoded in the objectives, we can also encode deterministic knowledge—such as a partial action model, known states or actions, background knowledge, static predicates, and model restrictions such as well-formedness~\cite{gjg:25}—as constraints directly into the MILP when such information is available.

\subsection{Integrating MILP with the Framework}
\begin{figure}[t]
  \centering
  \includegraphics[width=0.9\linewidth]{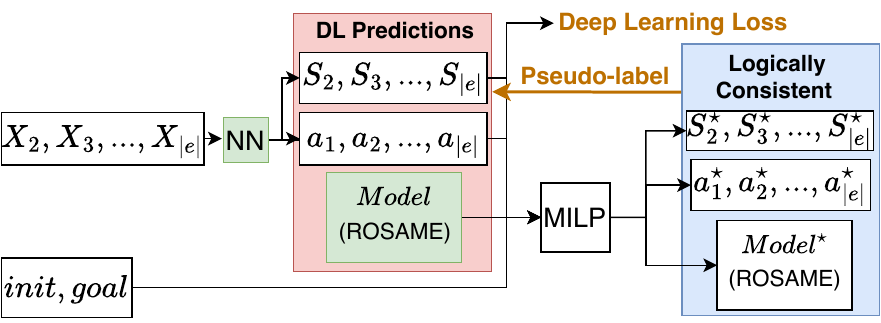}
  \caption{
    MILP module integrated as a plug-in to generate pseudo-labels from predicted states, actions and the action model, which guide further training.
  }
  \label{fig: dl_mip_int}
\end{figure}

Ideally, we would formulate a single MILP over all traces in the training dataset. However, this is infeasible in practice: the solving time grows rapidly with the number of variables, which scales with both the number of traces and the length of each trace. Instead, at regular intervals during training, we solve a MILP on a small, randomly selected subset of traces from the dataset.

We integrate the MILP solver as a plug-in module within the deep learning framework. The deep learning model is trained independently, and periodically we sample several visual traces from the dataset. For each sampled trace, we obtain its predicted states and predicted actions from the neural predictors, and we obtain the lifted action model from ROSAME. Using these predictions, we construct a MILP whose objective measures how closely a logically consistent solution aligns with them. Solving the MILP yields a globally consistent explanation of the sampled traces, from which we derive pseudo-labels that supervise subsequent training. \Figure~\ref{fig: dl_mip_int} illustrates this iterative interaction.

In this work, we treat action schema names and parameters as pure symbols, ignoring their semantic meaning. As a result, permuting schema names with identical signatures, or permuting parameters of the same type within a schema, produces another valid representation of the same underlying action model. Consequently, multiple equivalent solutions may exist due to this symmetry. The deep learning model may converge to one such variant, while the MILP may produce another. To align these representations, we enumerate all valid permutations of schema names and parameters for the MILP solution. For each permutation, we evaluate its agreement with the ROSAME action model according to \Equation~\ref{eq: rosame_pre}–\ref{eq: rosame_del}. We then select the permutation with the highest agreement and apply it to the MILP solution before extracting pseudo-labels.

\begin{table*}[th!]
\centering
\small
{
\begin{tabular}{|l|l|l|l|l|l|l|l|l|l|l|}
\hline
Domains                                                             & $|P_I|$ & $|A_I|$ & \# Steps & \# Fixes & TL(MILP){[}s{]} & Err & Agree & State Acc & Action Acc & t/ep{[}s{]} \\ \hline
\begin{tabular}[c]{@{}l@{}}Blocksworld (MNIST grid)\end{tabular}    & 36      & 50      & 3        & 4        & 60              & 0   & 0.977 & 97.81\%   & 85.33\%    & 2.50           \\ \hline
Gripper                                                             & 28      & 50      & 9        & 3        & 150             & 0   & 0.978 & 100\%     & 100\%      & 2.91           \\ \hline
Logistics                                                           & 72      & 196     & 9        & 3        & 150             & 0   & 0.983 & 99.89\%   & 99.56\%    & 10.64          \\ \hline
\begin{tabular}[c]{@{}l@{}}Blocksworld (Synthesized)\end{tabular}   & 36      & 50      & 3        & 3        & 60              & 0   & 0.976 & 99.29\%   & 88.67\%    & 1.51           \\ \hline
Hanoi                                                               & 55      & 120     & 5        & 3        & 90              & 0   & 0.940 & 98.55\%   & 81.40\%    & 4.77           \\ \hline
8-puzzle                                                            & 153     & 576     & 5        & 3        & 90              & 0   & 0.985 & 99.77\%   & 92.60\%    & 14.54          \\ \hline
\end{tabular}
}
\caption{Evaluation results across domains. TL: MILP time limit; t/ep: average training time per epoch.}
\label{tab: main-results}
\end{table*}

\subsection{MILP Pseudo-label Supervision}
The MILP solution provides pseudo-labels for three components of the neural model: the state predictor, the action predictor, and ROSAME.

\subsubsection{State supervision.}
For each time step $t$, the binary state vector $s_t^\star \in \{0,1\}^{|P_I|}$ is obtained from the MILP decision variables via $s_{t,i}^\star = 1$ iff \mipvar{hol}$[p_i,t] = 1$. This serves as the target for the probabilistic state prediction $\bm{ps}_t$. We supervise this prediction using binary cross-entropy.

\subsubsection{Action supervision.}
Since the MILP enforces that exactly one action is executed at each time step, we define $a_t^\star \in A_I$ as the (unique) action satisfying \mipvar{act[$a_t^\star$,t]} = 1. We replace the self-regularization target $a_t^\ast$ in \Section~\ref{sec: dl_framework}, with $a_t^\star$ as the cross-entropy target.

\subsubsection{Action model supervision.}
For each action schema $\alpha$ and parameter-bound predicate $\langle \rho, b_{\rho,\alpha} \rangle$, ROSAME predicts a probability distribution over four mutually exclusive cases: irrelevant (not used in the action model), add effect, precondition (but not delete effect), or delete effect. From the MILP decision variables \mipvar{pre}, \mipvar{add}, and \mipvar{del}, we determine the target case $c^\star \in \{1,2,3,4\}$. We then train ROSAME using the cross-entropy loss against $c^\star$, averaged over all pairs of action schemas and parameter-bound predicates.

\subsection{Pseudo-label Aging and Updating}
The quality of a MILP solution depends strongly on the accuracy of the neural predictions used to construct it. Early in training, these predictions may be noisy, and solving the MILP on a small subset of traces can yield solutions that do not generalize across the dataset. Moreover, MILP solutions obtained at different epochs may not be mutually consistent.

To account for this, we apply an exponential decay to each pseudo-label based on its age. If a pseudo-label is created at epoch $e_0$, then at epoch $e > e_0$ its loss contribution is weighted by $\psi^{\,e - e_0}$ with $\psi < 1$. This schedule ensures that older pseudo-labels gradually lose influence, while newer pseudo-labels—derived from increasingly accurate predictions—are emphasized. When a trace is sampled again during training, its pseudo-labels are recomputed and replace the previous ones.

Finally, the MILP objective itself is flexible: we may include all, some, or weighted combinations of its terms depending on our confidence in the corresponding neural predictions. This flexibility allows the MILP to emphasize reliable predictions while downweighting inconsistent ones, thereby providing more effective supervision for learning.

\section{Experiments}
We implement the deep learning framework in PyTorch 2.6 and use Gurobi 12.0.1 as the MILP solver. All experiments were conducted on a machine equipped with an AMD Ryzen 9 9900X CPU (12 logical cores), 24 GB of RAM, and a single NVIDIA RTX 4090 GPU.

\subsubsection{Visual trace dataset.}
We use the same dataset as \citet{xgt:24}. There are five domains—Blocksworld, Gripper, Logistics, Hanoi, and 8-puzzle—and two types of visualization: (1) digit and letter figures from the MNIST~\cite{d:12} and EMNIST~\cite{catv:17} datasets to represent objects and backgrounds, arranged in grids for Blocksworld, Gripper, and Logistics; and (2) synthesized images rendered with the PDDLGym library~\cite{sc:20} for Blocksworld, Hanoi, and 8-puzzle. Note that multiple distinct images may represent the same state. For Blocksworld (both visual representations), block piles may occupy different horizontal positions on the table; for Gripper, balls may appear in different positions within the same room; and for Logistics, objects may occupy different grid cells assigned to the same location.

The object sets for each domain are fixed across all traces: 5 blocks for Blocksworld (both visual representations); 6 balls, 2 grippers, and 2 rooms for Gripper; 6 packages, 2 cities (each with 2 locations, one of which is an airport), 2 trucks, and 2 airplanes for Logistics; 4 discs and 3 pegs for Hanoi; and 8 tiles on a $3\times3$ board for 8-puzzle.

For each trace, we sample a random state as the initial state using a problem generator, and generate a trace by random walk, uniformly selecting an applicable action at every step. For Blocksworld and 8-puzzle, the generator is provided by the PDDL-generators library~\cite{sth:22}; for the remaining domains, we implement problem generators ourselves, since the library does not support generating the range of random initial states we require.

\subsubsection{Training.}
We generate 2,000 traces per domain and split them into training and test sets (90:10). We train for 5,000 epochs using Adam with $\beta=(0.9,0.999)$, learning rate $10^{-4}$, and batch size 128. We set the prior-bias weight $\lambda=0.4$ and the supervision weight $\gamma=10$. We also apply data augmentation with image variations of the same state.

Training proceeds in two phases. We begin by training the deep learning model alone for 50 epochs, without MILP integration. At every epoch thereafter, we randomly sample a small set of 3--4 traces from the training set and formulate a corresponding MILP problem. Each MILP is solved with a time limit, and pseudo-labels are extracted from the best feasible solution, if any. These pseudo-labels are used to supervise the neural model, with their influence annealed over time using an exponential decay factor $\psi = 0.99$.

\subsubsection{Evaluation and results.}
To evaluate the learned action model, we follow \citet{xgt:24} in reporting the \emph{Error}, defined as the syntactic difference between the learned and ground-truth models under the best valid permutation of action schema names and parameters. We also report the maximum agreement score (normalized to $[0,1]$) between the learned model and the ground truth (as defined in \Section~\ref{sec:MIP}). For state prediction, we compute proposition-wise accuracy using a threshold of $0.5$. For action prediction, we report accuracy under the best-matching schema and parameter permutation identified during evaluation.

\Table~\ref{tab: main-results} summarizes the results. For each domain, we report results for the longest trace length we considered given the time available. We recover the ground-truth action model (up to permutation) and achieve high accuracy in both state and action prediction across all domains. In addition to evaluation metrics, we report the number of traces which the MILP aims to fix at each epoch, the MILP time limit, and the average training time per epoch. We give longer time limits to longer traces, since the MILP problems are larger. During training, we observe that the MILP often requires substantial time to fix traces in early epochs and may produce suboptimal corrections under the time limit. As training progresses, MILP problems become easier to solve, and the average time per epoch becomes much lower than the time limit.

In general, the problem becomes harder as visual traces become longer, for two reasons. First, longer visual traces provide weaker indirect supervision for intermediate states. Second, the complexity of solving the MILP grows rapidly with trace length, limiting how much correction the MILP can provide. In our experiments, agreement with the ground truth model consistently decreases as trace length increases. Interestingly, however, state and action accuracy sometimes increase for longer traces. Our hypothesis is that, since pseudo-labels are generated via MILP, longer traces yield more pseudo-supervision for both states and actions, which can counteract the increased learning difficulty.

\subsubsection{Ablation study.}
\begin{figure}[t]
  \centering
  \includegraphics[width=0.949\linewidth]{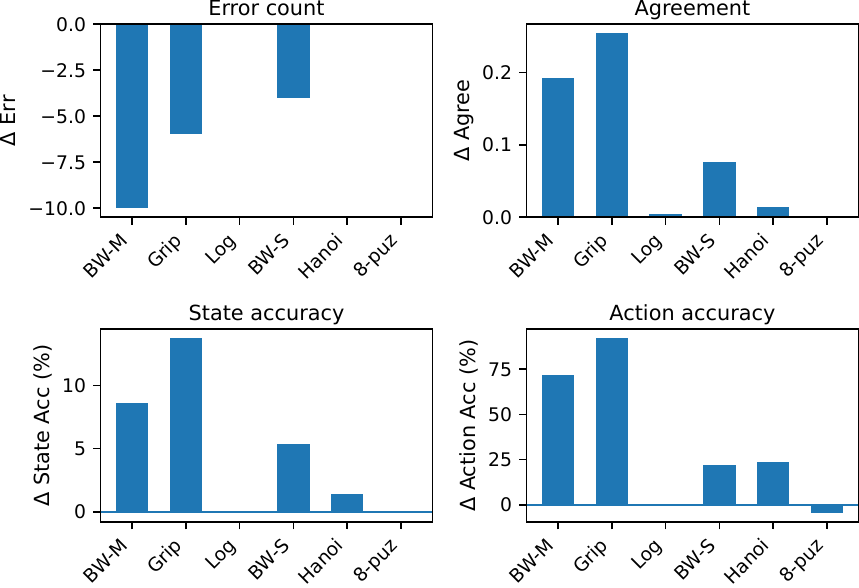}
  \caption{
    {Ablation of MILP correction. Bars show (with MILP) – (without MILP).}
  }
  \label{fig: mip_ablation}
\end{figure}
\begin{figure}[t]
    \centering
    \includegraphics[width=0.949\linewidth]{
    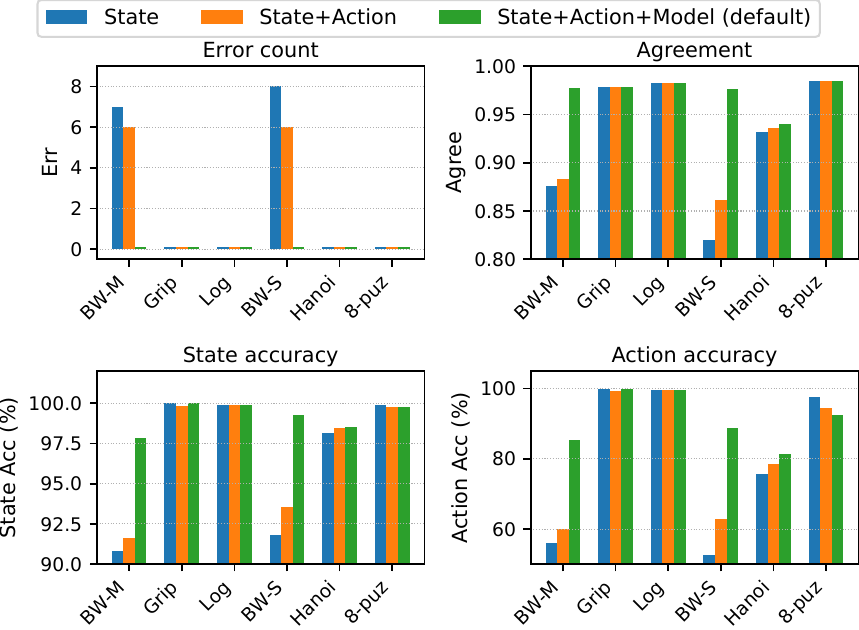}
    \caption{{Comparison among different MILP objectives}}
    \label{fig: mip_objectives}
\end{figure}
\Figure~\ref{fig: mip_ablation} compares learning outcomes with and without MILP. Each bar shows the change in performance obtained by adding MILP on top of the baseline trained \emph{without} MILP. Where needed, incorporating pseudo-label supervision from MILP prevents prediction collapse and helps training escape local optima, enabling convergence to a globally consistent solution. Including MILP reduces model error, improves agreement with the ground truth, and increases both state and action prediction accuracy. For Logistics and 8-puzzle, however, the baseline neural model already attains a correct solution without MILP.

The MILP objective incorporates terms corresponding to state predictions, action predictions, and the action model predicted by ROSAME. As discussed in \Section~\ref{sec:MIP}, this objective can be flexibly customized. \Figure~\ref{fig: mip_objectives} compares three different configurations of the MILP objective: (1) using only state predictions, (2) using both state and action predictions, and (3) using all terms. While several domains show comparable results across the three configurations, in Blocksworld (both visual representations) including all three terms significantly outperforms the other two. Allowing fewer terms in the objective increases correction flexibility but reduces fidelity to the neural model’s predictions; these effects compete and are difficult to balance. Identifying a more principled way to balance these terms remains future work.

\section{Conclusion, Related and Future Work}
Our work contributes to the broader problem of automatically constructing planning models from data. Beyond action model learning, a fully autonomous pipeline also involves defining or inventing a symbolic vocabulary that is sufficient for planning~\cite{kkl:14, kkl:18, lktetshe:25, aksllk:25}, abstracting available data into a symbolic representation~\cite{jrk:22, vms:22, lktetshe:25}, repairing or reconciling inconsistencies in the learned models~\cite{gfgf:23, lgb:23}, and integrating the resulting models with executable low-level skills~\cite{cstlk:22, sns:25}. In this work, we focus on the intersection of symbolic abstraction and action model learning in the context of visual observations. In particular, we address errors and uncertainty introduced during abstraction by leveraging constraints induced by action model and trace consistency. Works that focus on other components of this pipeline are complementary but outside the scope of this paper. Ultimately, a complete solution will require a unified framework that jointly addresses all components.

In this paper, we propose a framework for learning lifted action models from visual traces without action supervision. A closely related line of work is Latplan~\cite{af:17, af:18, ak:19, am:20, akfm:22}, an unsupervised framework that learns \textit{latent, propositional} action models from state image pairs. Latplan does not rely on a known symbolic vocabulary or fully specified initial/final states like we do. Meanwhile, its learned model is uninterpretable, cannot generalize to instances with different object counts or visual styles, and cannot be used to plan without images. In contrast, we assume access to the domain’s predicates and action schema signatures, which allow us to leverage ROSAME~\cite{xgt:24} to learn lifted, interpretable action models that generalize across instances. They also enable us to create a simpler and more transparent framework than Latplan.

\citet{bg:20} investigate a different but related setting, where lifted action models—as well as predicates and action schemas—are learned from a labeled graph representing the state space structure for small instances. \citet{rbrg:21} extend this framework to handle partial knowledge of the graph and a limited form of noise that obscures the identity of the target state for certain transitions. These approaches rely on different assumptions and offer complementary strengths.

Our MILP model is also related to model repair methods such as FAMA~\citep{aco:19}, which learns or repairs lifted action models—sometimes using only initial and final states—via compilation into classical planning with context-dependent effects. A key difference is that our approach incorporates \emph{probabilistic} knowledge about states, actions, and the action model, whereas FAMA requires deterministic knowledge (which may also be incorporated in our MILP). While it may be possible to simulate our MILP within the FAMA compilation by using context-dependent action costs, few optimal planners support this feature~\citep{gkm:16, igh:19}, and those that do remain less competitive than mature MILP solvers.

The scalability of our approach is currently bounded by the complexity of solving the MILP. Future work will explore more scalable constraint-solving or approximation techniques that preserve logical structure while reducing computational cost, enabling correction over more and longer traces. Another important direction is to remove the assumption that predicates and action schemas are provided.

\section*{Acknowledgments}
This work was funded by the Australian Research Council (ARC) under the Discovery Project grant DP220103815 and by the Artificial and Natural Intelligence Toulouse Institute (ANITI) under the grant agreement ANR-23-IACL-0002. We also thank the the anonymous reviewers for their helpful suggestions.

\bibliography{aaai2026}

\appendix
\clearpage
\section{ROSAME Implementation Details}
In this section, we provide technical details about the implementation of ROSAME~\cite{xgt:24}. Note however that they are not needed to understand the paper. Given a set of predicate symbols $\mathcal{P}$ and action symbols $\mathcal{A}$, ROSAME assumes that any valid action model must satisfy the following two constraints:
\begin{itemize}
    \item add effects and preconditions cannot intersect, i.e., $\Add(\alpha) \cap \Pre(\alpha) = \emptyset$;
    \item only preconditions can be deleted, i.e., $\Del(\alpha) \subseteq \Pre(\alpha)$.
\end{itemize}
Based on these constraints, ROSAME enumerates all valid relationships between an action schema $\alpha$ and a parameter-bound predicate $\langle \rho, b_{\rho, \alpha} \rangle$. There are four mutually exclusive cases:
\begin{itemize}
    \item \textbf{Case 1:} $\langle \rho, b_{\rho, \alpha} \rangle$ is not involved in the action model of $\alpha$.
    \item \textbf{Case 2:} $\langle \rho, b_{\rho, \alpha} \rangle$ is an add effect only.
    \item \textbf{Case 3:} $\langle \rho, b_{\rho, \alpha} \rangle$ is a precondition only.
    \item \textbf{Case 4:} $\langle \rho, b_{\rho, \alpha} \rangle$ is both a precondition and a delete effect (but not an add effect).
\end{itemize}
ROSAME frames the task of learning action models as a classification problem: for each pair $(\alpha, \langle \rho, b_{\rho, \alpha} \rangle)$, it predicts which of the four cases applies. To do this, ROSAME trains a deep learning classifier for each action schema $\alpha$, where the inputs are fixed latent vectors—one for each parameter-bound predicate $\langle \rho, b_{\rho, \alpha} \rangle$—and the outputs are probability distributions over the four cases. 

Let $\overrightarrow{pr}_{\alpha, \langle \rho, b_{\rho, \alpha} \rangle}$ denote the predicted probability vector for the four cases. These probabilities are then decoded into the lifted precondition and effect indicators via:
\begin{align*}
  pre_{\alpha, \langle \rho, b_{\rho, \alpha} \rangle} &= \overrightarrow{pr}_{\alpha, \langle \rho, b_{\rho, \alpha} \rangle} \cdot (0, 0, 1, 1) \\
  add_{\alpha, \langle \rho, b_{\rho, \alpha} \rangle} &= \overrightarrow{pr}_{\alpha, \langle \rho, b_{\rho, \alpha} \rangle} \cdot (0, 1, 0, 0) \\
  del_{\alpha, \langle \rho, b_{\rho, \alpha} \rangle} &= \overrightarrow{pr}_{\alpha, \langle \rho, b_{\rho, \alpha} \rangle} \cdot (0, 0, 0, 1) 
\end{align*}

These expressions compute the expected value for each label by summing over the compatible cases. For instance, the probability that $\langle \rho, b_{\rho, \alpha} \rangle$ is a precondition is the sum of the probabilities assigned to cases 3 and 4. 

By aggregating the outputs over all parameter-bound predicates, ROSAME constructs the full lifted precondition and effect vectors $pre_{\alpha}$, $add_{\alpha}$, and $del_{\alpha}$ for each action schema $\alpha$.

\section{Visual Trace Dataset}
\begin{figure*}
  \centering
  \includegraphics[width=0.4\textwidth]{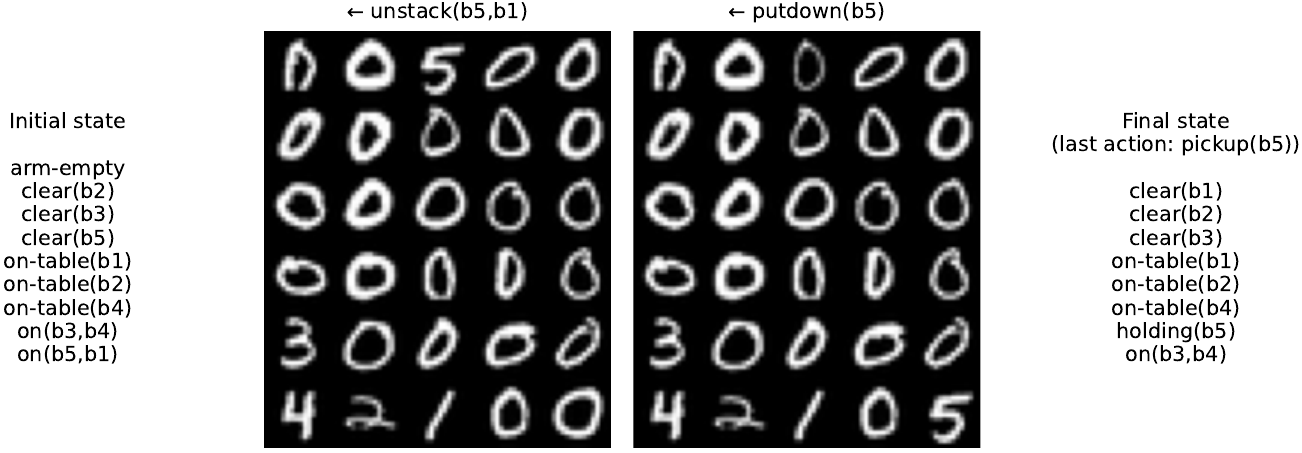}
  \caption{A 3-step visual trace example for Blocksworld (MNIST grid).}
  \label{fig: blocksworld_example}
\end{figure*}

\begin{figure*}
  \centering
  \includegraphics[width=\textwidth]{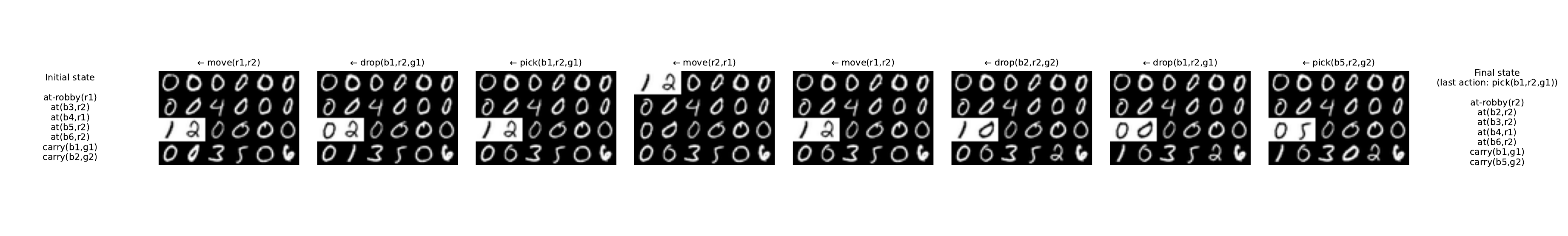}
  \caption{A 9-step visual trace example for Gripper.}
  \label{fig: gripper_example}
\end{figure*}

\begin{figure*}
  \centering
  \includegraphics[width=\textwidth]{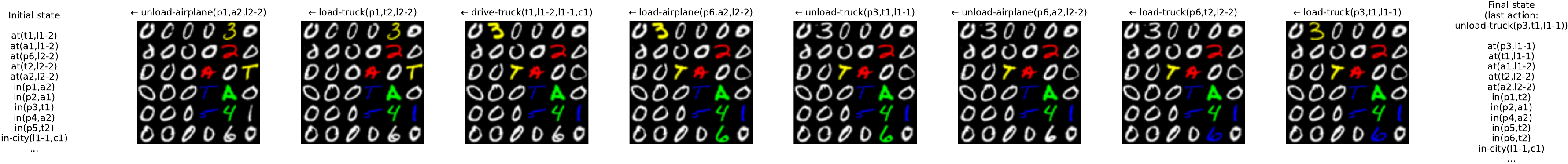}
  \caption{A 9-step visual trace example for Logistics.}
  \label{fig: logistics_example}
\end{figure*}

\begin{figure*}
  \centering
  \includegraphics[width=0.4\textwidth]{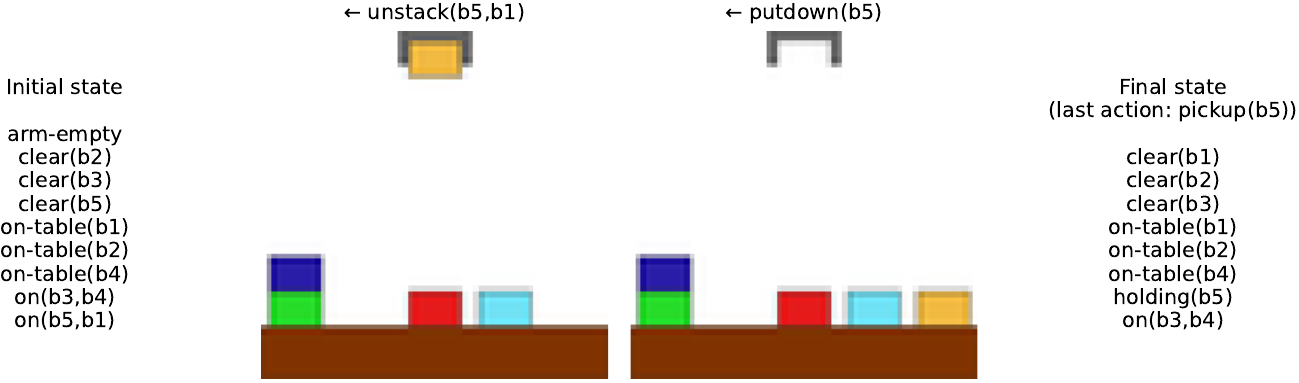}
  \caption{A 3-step visual trace example for Blocksworld (Synthesized).}
  \label{fig: blocksworld_pddlgym_example}
\end{figure*}

\begin{figure*}
  \centering
  \includegraphics[width=0.4\textwidth]{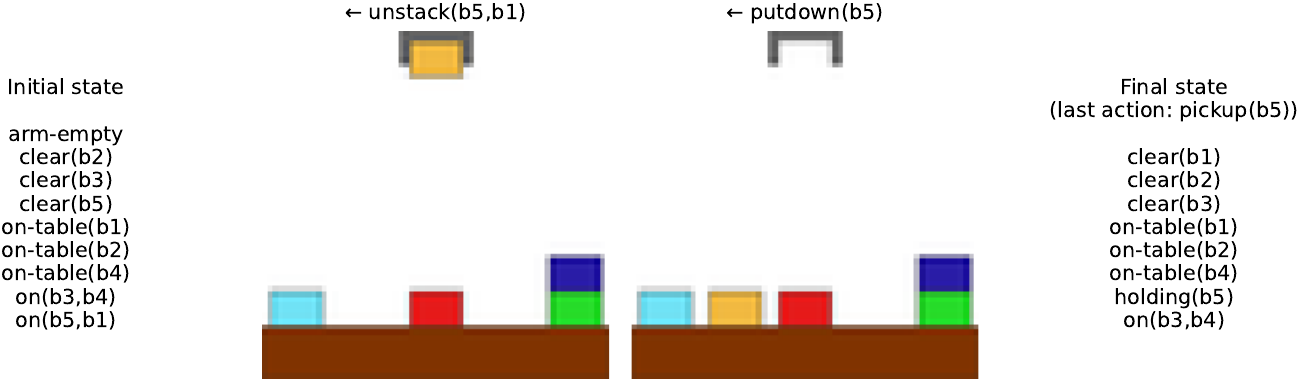}
  \caption{An alternative visual trace corresponding to the same underlying state trace as \Figure~\ref{fig: blocksworld_pddlgym_example}.}
  \label{fig: blocksworld_alter_example}
\end{figure*}

\begin{figure*}
  \centering
  \includegraphics[width=0.6\textwidth]{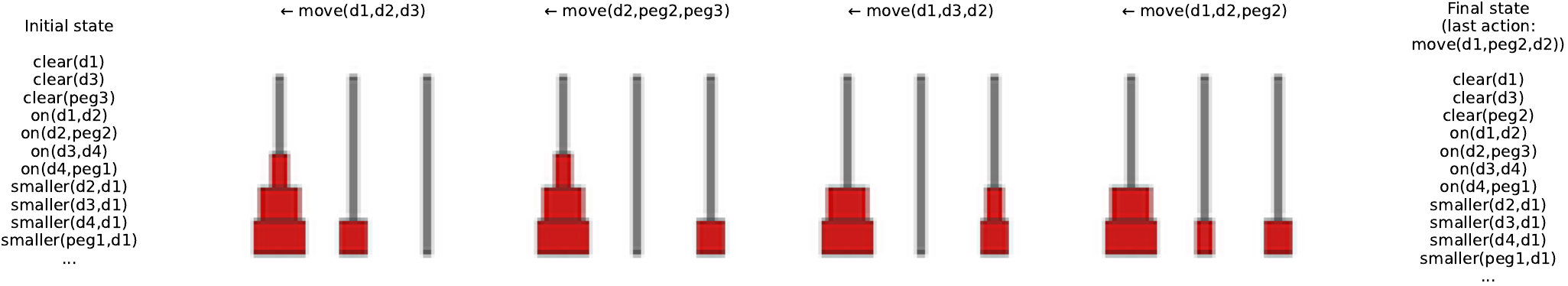}
  \caption{A 5-step visual trace example for Hanoi.}
  \label{fig: hanoi_example}
\end{figure*}

\begin{figure*}
  \centering
  \includegraphics[width=0.6\textwidth]{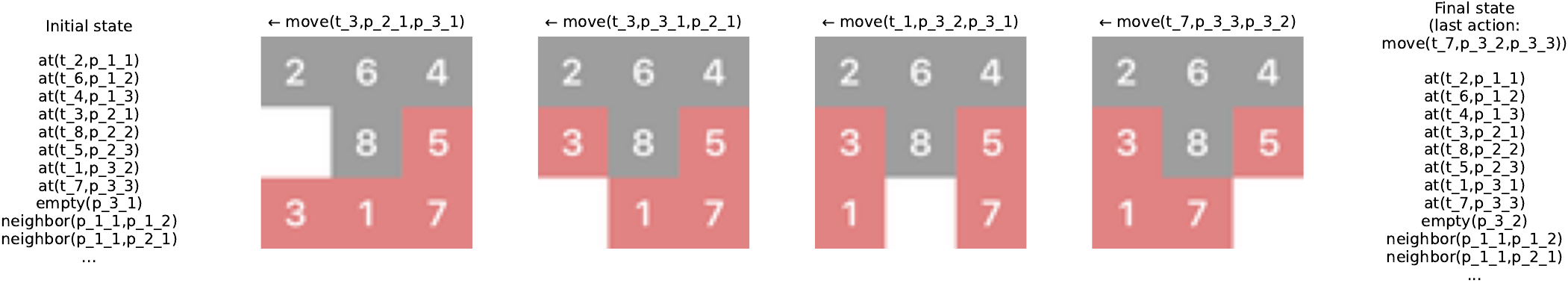}
  \caption{A 5-step visual trace example for 8-puzzle.}
  \label{fig: 8-puzzle_example}
\end{figure*}

\begin{table*}[th!]
\centering
{\small
\begin{tabular}{|l|l|l|l|l|l|l|l|l|l|l|}
\hline
Domains                                                             & $|P_I|$              & $|A_I|$              & \# Steps & \# Fixes           & TL(MILP){[}s{]} & Err & Agree & State Acc & Action Acc & t/epoch{[}s{]} \\ \hline
\begin{tabular}[c]{@{}l@{}}Blocksworld\\ (MNIST grid)\end{tabular}  & 36                   & 50                   & 3        & 4                  & 60              & 0   & 0.977 & 97.81\%   & 85.33\%    & 2.50           \\ \hline
\multirow{4}{*}{Gripper}                                            & \multirow{4}{*}{28}  & \multirow{4}{*}{50}  & 3        & \multirow{4}{*}{3} & 60              & 0   & 0.990 & 97.07\%   & 83.33\%    & 1.07           \\ \cline{4-4} \cline{6-11} 
                                                                    &                      &                      & 5        &                    & 90              & 0   & 0.984 & 100\%     & 100\%      & 1.60           \\ \cline{4-4} \cline{6-11} 
                                                                    &                      &                      & 7        &                    & 120             & 0   & 0.980 & 100\%     & 100\%      & 2.03           \\ \cline{4-4} \cline{6-11} 
                                                                    &                      &                      & 9        &                    & 150             & 0   & 0.978 & 100\%     & 100\%      & 2.91           \\ \hline
\multirow{4}{*}{Logistics}                                          & \multirow{4}{*}{72}  & \multirow{4}{*}{196} & 3        & \multirow{4}{*}{3} & 60              & 0   & 0.992 & 99.28\%   & 98.33\%    & 2.67           \\ \cline{4-4} \cline{6-11} 
                                                                    &                      &                      & 5        &                    & 90              & 0   & 0.987 & 99.62\%   & 97.80\%    & 4.42           \\ \cline{4-4} \cline{6-11} 
                                                                    &                      &                      & 7        &                    & 120             & 0   & 0.984 & 99.83\%   & 99.71\%    & 6.43           \\ \cline{4-4} \cline{6-11} 
                                                                    &                      &                      & 9        &                    & 150             & 0   & 0.983 & 99.89\%   & 99.56\%    & 10.64          \\ \hline
\begin{tabular}[c]{@{}l@{}}Blocksworld\\ (Synthesized)\end{tabular} & 36                   & 50                   & 3        & 3                  & 60              & 0   & 0.976 & 99.29\%   & 88.67\%    & 1.51           \\ \hline
\multirow{2}{*}{Hanoi}                                              & \multirow{2}{*}{55}  & \multirow{2}{*}{120} & 3        & \multirow{2}{*}{3} & 60              & 0   & 0.942 & 96.42\%   & 62.00\%    & 2.35           \\ \cline{4-4} \cline{6-11} 
                                                                    &                      &                      & 5        &                    & 90              & 0   & 0.940 & 98.55\%   & 81.40\%    & 4.77           \\ \hline
\multirow{2}{*}{8-puzzle}                                           & \multirow{2}{*}{153} & \multirow{2}{*}{576} & 3        & \multirow{2}{*}{3} & 60              & 0   & 0.990 & 99.85\%   & 96.33\%    & 8.34           \\ \cline{4-4} \cline{6-11} 
                                                                    &                      &                      & 5        &                    & 90              & 0   & 0.985 & 99.77\%   & 92.60\%    & 14.54          \\ \hline
\end{tabular}
}
\caption{Full evaluation results}
\label{tab: full results}
\end{table*}

\begin{table*}[th!]
\centering
{\small
\begin{tabular}{|ll|l|l|l|l|}
\hline
\multicolumn{2}{|l|}{Domain}                                                                                           & Err & Agree & \begin{tabular}[c]{@{}l@{}}State\\ Acc\end{tabular} & \begin{tabular}[c]{@{}l@{}}Action \\ Acc\end{tabular} \\ \hline
\multicolumn{1}{|l|}{\multirow{2}{*}{\begin{tabular}[c]{@{}l@{}}Blocksworld\\ (MNIST grid)\end{tabular}}}  & w/o MILP  & 10  & 0.784 & 89.22\%                                             & 13.67\%                                               \\ \cline{2-6} 
\multicolumn{1}{|l|}{}                                                                                     & with MILP & 0   & 0.977 & 97.81\%                                             & 85.33\%                                               \\ \hline
\multicolumn{1}{|l|}{\multirow{2}{*}{Gripper}}                                                             & w/o MILP  & 6   & 0.724 & 86.22\%                                             & 7.60\%                                                 \\ \cline{2-6} 
\multicolumn{1}{|l|}{}                                                                                     & with MILP & 0   & 0.978 & 100\%                                               & 100\%                                                 \\ \hline
\multicolumn{1}{|l|}{\multirow{2}{*}{Logistics}}                                                           & w/o MILP  & 0   & 0.979 & 99.93\%                                             & 99.67\%                                               \\ \cline{2-6} 
\multicolumn{1}{|l|}{}                                                                                     & with MILP & 0   & 0.983 & 99.89\%                                             & 99.56\%                                               \\ \hline
\multicolumn{1}{|l|}{\multirow{2}{*}{\begin{tabular}[c]{@{}l@{}}Blocksworld\\ (Synthesized)\end{tabular}}} & w/o MILP  & 4   & 0.899 & 93.90\%                                             & 66.67\%                                               \\ \cline{2-6} 
\multicolumn{1}{|l|}{}                                                                                     & with MILP & 0   & 0.976 & 99.29\%                                             & 88.67\%                                               \\ \hline
\multicolumn{1}{|l|}{\multirow{2}{*}{Hanoi}}                                                               & w/o MILP  & 0   & 0.926 & 97.15\%                                             & 57.60\%                                               \\ \cline{2-6} 
\multicolumn{1}{|l|}{}                                                                                     & with MILP & 0   & 0.940 & 98.55\%                                             & 81.40\%                                               \\ \hline
\multicolumn{1}{|l|}{\multirow{2}{*}{8-puzzle}}                                                            & w/o MILP  & 0   & 0.985 & 99.90\%                                             & 97.40\%                                               \\ \cline{2-6} 
\multicolumn{1}{|l|}{}                                                                                     & with MILP & 0   & 0.985 & 99.77\%                                             & 92.60\%                                               \\ \hline
\end{tabular}
}
\caption{Comparison between with and without MILP}
\label{tab: ablation mip}
\end{table*}

\begin{table*}[th!]
\centering
{\small
\begin{tabular}{|l|l|l|l|l|l|}
\hline
Domains                                                                              & MILP Objective       & Err & Agree & \begin{tabular}[c]{@{}l@{}}State\\ Acc\end{tabular} & \begin{tabular}[c]{@{}l@{}}Action\\ Acc\end{tabular} \\ \hline
\multirow{3}{*}{\begin{tabular}[c]{@{}l@{}}Blocksworld\\ (MNIST Grid)\end{tabular}}  & State                & 7   & 0.876 & 90.79\%                                             & 56.00\%                                              \\ \cline{2-6} 
                                                                                     & State, Action        & 6   & 0.883 & 91.62\%                                             & 60.00\%                                              \\ \cline{2-6} 
                                                                                     & State, Action, Model & 0   & 0.977 & 97.81\%                                             & 85.33\%                                              \\ \hline
\multirow{3}{*}{Gripper}                                                             & State                & 0   & 0.978 & 100\%                                               & 100\%                                                \\ \cline{2-6} 
                                                                                     & State, Action        & 0   & 0.978 & 99.85\%                                             & 99.22\%                                              \\ \cline{2-6} 
                                                                                     & State, Action, Model & 0   & 0.978 & 100\%                                               & 100\%                                                \\ \hline
\multirow{3}{*}{Logistics}                                                           & State                & 0   & 0.983 & 99.86\%                                             & 99.67\%                                              \\ \cline{2-6} 
                                                                                     & State, Action        & 0   & 0.983 & 99.91\%                                             & 99.56\%                                              \\ \cline{2-6} 
                                                                                     & State, Action, Model & 0   & 0.983 & 99.89\%                                             & 99.56\%                                              \\ \hline
\multirow{3}{*}{\begin{tabular}[c]{@{}l@{}}Blocksworld\\ (Synthesized)\end{tabular}} & State                & 8   & 0.820 & 91.79\%                                             & 52.67\%                                              \\ \cline{2-6} 
                                                                                     & State, Action        & 6   & 0.861 & 93.54\%                                             & 63.00\%                                              \\ \cline{2-6} 
                                                                                     & State, Action, Model & 0   & 0.976 & 99.29\%                                             & 88.67\%                                              \\ \hline
\multirow{3}{*}{Hanoi}                                                               & State                & 0   & 0.932 & 98.18\%                                             & 75.60\%                                              \\ \cline{2-6} 
                                                                                     & State, Action        & 0   & 0.936 & 98.43\%                                             & 78.40\%                                              \\ \cline{2-6} 
                                                                                     & State, Action, Model & 0   & 0.940 & 98.55\%                                             & 81.40\%                                              \\ \hline
\multirow{3}{*}{8-puzzle}                                                            & State                & 0   & 0.985 & 99.90\%                                             & 97.60\%                                              \\ \cline{2-6} 
                                                                                     & State, Action        & 0   & 0.985 & 99.79\%                                             & 94.40\%                                              \\ \cline{2-6} 
                                                                                     & State, Action, Model & 0   & 0.985 & 99.77\%                                             & 92.60\%                                              \\ \hline
\end{tabular}
}
\caption{Comparison among different MILP objectives}
\label{tab: mip objectives}
\end{table*}

\Figure~\ref{fig: blocksworld_example}–\ref{fig: 8-puzzle_example} present example visual traces for the domains and visual representations we use in our experiments. In each example, we show the fully observed initial and final states, the intermediate state images, and the ground-truth actions executed (which we do not provide to the model during training). We now detail how we implement trace rendering.

\subsubsection{Blocksworld (MNIST grid)}
In Blocksworld, we use an object set of five blocks. To render state images with MNIST digits, for each trace we select random figures for digits 1–5 to represent the five blocks. The environment is organized as a $6\times5$ grid of cells. If a block is on the table, we randomly place it in one of five grid cells in the bottom row. If a block is held by the arm, we place it in the middle cell of the top row. If it is on top of another block, its position is the cell above its supporting block. We fill the backgrounds with randomly selected figures for digit 0. When we progress through the trace, if the executed action is a $\mathrm{pickup}$ action, we move the block to the middle cell of the top row and fill the cell it leaves with a random figure for digit 0. If the executed action is a $\mathrm{putdown}$ action, we randomly select an unoccupied cell in the bottom row and move the block there, and fill the cell it leaves (which is the middle cell of the top row) with a random figure for digit 0. $\mathrm{stack}$ and $\mathrm{unstack}$ are implemented similarly, except that the origin and the destination become the cell above the supporting block. Cells that are not affected by the executed action remain unchanged.

\subsubsection{Gripper}
In Gripper, we use an object set of six balls, two grippers, and two rooms. To render state images, for each trace we select random figures for digits 1–6 to represent the six balls. Each room is represented as a $2\times6$ grid of cells, and the two rooms are stacked vertically. If a ball is in a room, we randomly place it in one of six grid cells in the bottom row of that room. The two grippers are rendered as two cells with their background and foreground colors flipped. If the grippers are at a room, we place them at the two top-left cells of that room. If a gripper holds a ball, we fill its cell with the figure for that ball. Otherwise, we fill it with a random figure for digit 0. We also fill the backgrounds with randomly selected figures for digit 0. When we progress through the trace, if the executed action is a $\mathrm{pick}$ action, we move the ball to the cell of the gripper, flip its color, and fill the cell it leaves with a random figure for digit 0. If it is a $\mathrm{drop}$ action, we move the ball from the cell of the gripper to a randomly selected, unoccupied bottom-row cell of the room, and flip its color back. If it is a $\mathrm{move}$ action, we move the two grippers to the top-left of the other room and fill the cells they leave with two random figures for digit 0 with normal color.

\subsubsection{Logistics}
In Logistics, we use an object set of six packages, two cities (each with two locations, one of which is an airport), two trucks (one in each city), and two airplanes. To render state images, for each trace we select random figures for digits 1–6 to represent the six packages, two random figures for letter~T to represent the trucks, and two random figures for letter~A to represent the airplanes. We use four different colors to distinguish the two trucks and two airplanes. Each location is organized as a $3\times3$ grid of cells, and we stack the two locations in each city horizontally (the left one is the normal location, the right one the airport), and then stack the two cities vertically. If a package, truck, or airplane is at a location, we randomly place it in one of the cells assigned to that location. If a package is in a truck or airplane, we find the location of the vehicle and randomly place the package in one of the cells assigned to that location and color the figure with the vehicle’s assigned color. When we progress through the trace, if the executed action is a $\mathrm{load\mbox{-}truck}$ or $\mathrm{load\mbox{-}airplane}$ action, we color the package with the color assigned to the vehicle. If it is an $\mathrm{unload\mbox{-}truck}$ or $\mathrm{unload\mbox{-}airplane}$ action, we color the package back to black. If it is a $\mathrm{drive\mbox{-}truck}$ or $\mathrm{fly\mbox{-}airplane}$ action, we move the vehicle, together with all packages inside it, by placing them randomly in the cells assigned to the target location, and fill the cells they leave with random figures for digit~0.

\subsubsection{Blocksworld (Synthesized)}
We rely on the renderer implemented in the PDDLGym library~\citep{sc:20} to generate synthesized state images in Blocksworld. The object set has five blocks, and there are five slots on the table. When a block is on the table, or when it is put down, it is placed randomly in one of the empty slots. As a result, multiple visual traces may correspond to the same underlying state trace. \Figure~\ref{fig: blocksworld_pddlgym_example} and \Figure~\ref{fig: blocksworld_alter_example} show two visual traces that correspond to the same underlying state trace.

\section{Experimental Results}
\Table~\ref{tab: full results} presents the full evaluation results across all domains. In these experiments, we vary the trace length from 3 to the maximum considered in the paper for each given domain. \Table~\ref{tab: ablation mip} compares learning outcomes with and without the MILP module (corresponding to \Figure~5 in the main paper). \Table~\ref{tab: mip objectives} shows the results for different configurations of the MILP objective (corresponding to \Figure~6 in the main paper).

\section{Examples of Learned Action Models}
\Figure~\ref{fig: blocksworld_pddl}--\ref{fig: 8-puzzle_pddl} present examples of the learned action models. As discussed in \Section~\ref{sec:MIP}, we consider permutations of schema names with identical signatures, as well as permutations of parameters of the same type within a schema, to be equivalent to the ground truth.

\begin{figure*}[t]
\small
\begin{verbatim}
(define (domain blocks)
        (:requirements :strips :typing)
        (:types object)
        (:predicates
            (arm-empty )
            (clear ?a - object)
            (on-table ?a - object)
            (holding ?a - object)
            (on ?a - object ?b - object))
        
        (:action pickup
            :parameters (?a - object)
            :precondition (and (holding a))
            :effect (and (arm-empty) (clear a) (on-table a) (not (holding a)))
        )
        

        (:action putdown
            :parameters (?a - object)
            :precondition (and (arm-empty) (clear a) (on-table a))
            :effect (and (not (arm-empty)) (not (clear a)) (not (on-table a)) (holding a))
        )
        

        (:action stack
            :parameters (?a - object ?b - object)
            :precondition (and (clear b) (holding a))
            :effect (and (arm-empty) (clear a) (not (clear b)) (not (holding a)) (on a b))
        )
        

        (:action unstack
            :parameters (?a - object ?b - object)
            :precondition (and (arm-empty) (clear b) (on b a))
            :effect (and (not (arm-empty)) (clear a) (not (clear b)) (holding b)
                    (not (on b a)))
        )
        
)
\end{verbatim}
\caption{Learned model for the Blocksworld domain.}
\label{fig: blocksworld_pddl}
\end{figure*}

\begin{figure*}[t]
\small
\begin{verbatim}
(define (domain gripper)
        (:requirements :strips :typing)
        (:types room ball gripper - object)
        (:predicates
            (at-robby ?a - room)
            (at ?a - ball ?b - room)
            (free ?a - gripper)
            (carry ?a - ball ?b - gripper))
        
        (:action move
            :parameters (?a - room ?b - room)
            :precondition (and (at-robby b))
            :effect (and (at-robby a) (not (at-robby b)))
        )
        

        (:action pick
            :parameters (?a - ball ?b - gripper ?c - room)
            :precondition (and (at-robby c) (carry a b))
            :effect (and (at a c) (free b) (not (carry a b)))
        )
        

        (:action drop
            :parameters (?a - ball ?b - gripper ?c - room)
            :precondition (and (at-robby c) (at a c) (free b))
            :effect (and (not (at a c)) (not (free b)) (carry a b))
        )
        
)
\end{verbatim}
\caption{Learned model for the Gripper domain.}
\label{fig: gripper_pddl}
\end{figure*}

\begin{figure*}[t]
\small
\begin{verbatim}
(define (domain logistics)
        (:requirements :strips :typing)
        (:types
            movable location city - object
            obj transport - movable
            truck airplane - transport
            airport - location)
        (:predicates
            (at ?a - location ?b - movable)
            (in ?a - obj ?b - transport)
            (in-city ?a - city ?b - location))
        
        (:action load-truck
            :parameters (?a - location ?b - obj ?c - truck)
            :precondition (and (at a b) (at a c))
            :effect (and (not (at a b)) (in b c))
        )
        

        (:action load-airplane
            :parameters (?a - airplane ?b - airport ?c - obj)
            :precondition (and (at b a) (at b c))
            :effect (and (not (at b c)) (in c a))
        )
        

        (:action unload-truck
            :parameters (?a - location ?b - obj ?c - truck)
            :precondition (and (at a c) (in b c))
            :effect (and (at a b) (not (in b c)))
        )
        

        (:action unload-airplane
            :parameters (?a - airplane ?b - airport ?c - obj)
            :precondition (and (at b a) (in c a))
            :effect (and (at b c) (not (in c a)))
        )
        

        (:action drive-truck
            :parameters (?a - city ?b - location ?c - location ?d - truck)
            :precondition (and (at b d) (in-city a b) (in-city a c))
            :effect (and (not (at b d)) (at c d))
        )
        

        (:action fly-airplane
            :parameters (?a - airplane ?b - airport ?c - airport)
            :precondition (and (at b a))
            :effect (and (not (at b a)) (at c a))
        )
        
)
\end{verbatim}
\caption{Learned model for the Logistics domain.}
\label{fig: logistics_pddl}
\end{figure*}

\begin{figure*}[t]
\small
\begin{verbatim}
(define (domain hanoi)
        (:requirements :strips :typing)
        (:types
            disc - object)
        (:predicates
            (clear ?a - object)
            (on ?a - disc ?b - object)
            (smaller ?a - disc ?b - object))
        
        (:action move
            :parameters (?a - disc ?b - object ?c - object)
            :precondition (and (clear a) (clear b) (on a c) (smaller a b) (smaller a c))
            :effect (and (not (clear b)) (clear c) (on a b) (not (on a c)))
        )
        
)
\end{verbatim}
\caption{Learned model for the Hanoi domain.}
\label{fig: hanoi_pddl}
\end{figure*}

\begin{figure*}[t]
\small
\begin{verbatim}
(define (domain 8-puzzle)
        (:requirements :strips :typing)
        (:types
            tile position - object)
        (:predicates
            (at ?a - position ?b - tile)
            (blank ?a - position)
            (neighbor ?a - position ?b - position))
        
        (:action move
            :parameters (?a - position ?b - position ?c - tile)
            :precondition (and (at a c) (blank b) (neighbor a b) (neighbor b a))
            :effect (and (not (at a c)) (at b c) (blank a) (not (blank b)))
        )
        
)
\end{verbatim}
\caption{Learned model for the 8-puzzle domain.}
\label{fig: 8-puzzle_pddl}
\end{figure*}

\end{document}